\documentclass[review]{elsarticle}
\usepackage{lineno,hyperref}
\usepackage{url}
\usepackage{graphicx}
\usepackage{algorithm}
\usepackage{algorithmic}
\usepackage{caption2}
\usepackage{subfigure}
\usepackage{float}
\usepackage{booktabs}
\usepackage{threeparttable}
\usepackage{multirow}
\usepackage{multicol}
\usepackage{arydshln}
\usepackage{amsfonts}
\usepackage{amsmath}
\usepackage{color}
\DeclareMathOperator{\sign}{sign}
\journal{Journal of \LaTeX\ Templates}
 
\begin{document}

\begin{frontmatter}

    \title{Adversarial symmetric GANs: bridging adversarial samples and adversarial networks}

    \author{Faqiang Liu\fnref{ec}}
    \author{Mingkun Xu\fnref{ec}}
    \author{Guoqi Li}
    \author{Jing Pei}
    \author{Luping Shi}
    \author{Rong Zhao\corref{mycorrespondingauthor}}
    \address{Department of Precision Instrument,Tsinghua University,Beijing,100084,China\\
        Center for Brain Inspired Computing Research,Tsinghua University,Beijing,100084,China\\
        Beijing Innovation Center for Future Chip,Beijing,100084,China
    }
    \fntext[ec]{Equal contribution}
    \cortext[mycorrespondingauthor]{Corresponding author:r\_zhao@mail.tsinghua.edu.cn}

    \begin{abstract}
        Generative adversarial networks have achieved remarkable performance on various tasks but suffer from training instability. Despite many training strategies proposed to improve training stability, this issue remains as a challenge. In this paper, we investigate the training instability from the perspective of adversarial samples and reveal that adversarial training on fake samples is implemented in vanilla GANs, but adversarial training on real samples has long been overlooked. Consequently, the discriminator is extremely vulnerable to adversarial perturbation and the gradient given by the discriminator contains non-informative adversarial noises, which hinders the generator from catching the pattern of real samples. Here, we develop adversarial symmetric GANs (AS-GANs) that incorporate adversarial training of the discriminator on real samples into vanilla GANs, making adversarial training symmetrical. The discriminator is therefore more robust and provides more informative gradient with less adversarial noise, thereby stabilizing training and accelerating convergence. The effectiveness of the AS-GANs is verified on image generation on CIFAR-10, CIFAR-100, CelebA, and LSUN with varied network architectures. Not only the training is more stabilized, but the FID scores of generated samples are consistently improved by a large margin compared to the baseline. Theoretical analysis is also conducted to explain why AS-GAN can improve training. The bridging of adversarial samples and adversarial networks provides a new approach to further develop adversarial networks.
    \end{abstract}


\end{frontmatter}

\section{Introduction}
Generative adversarial networks (GANs) have been applied successfully in various research fields such as natural image modeling \cite{Radford2015Unsupervised}, image translation \cite{Isola2016Image,Zhu2017Unpaired}, cross-modal image generation \cite{Dash2017TAC}, image super-resolution \cite{Ledig2016Photo}, semi-supervised learning \cite{Odena2016Semi} and sequential data modeling \cite{Mogren2016C,Yu2016SeqGAN}. Different from explicit density estimation based models \cite{kingma2014semi,Oord2016Pixel,hinton2012a}, GANs are implicit generative models with two neural networks playing min-max game to find a map from random noise to the target distribution, in which the generator tries to generate fake samples to fool the discriminator and the discriminator tries to distinguish them from real samples \cite{Goodfellow2014Generative}. In the original GANs, the optimal discriminator measures the Jensen-Shannon divergence between the real data distribution and the generated distribution. The discrepancy measure can be generalized to f-divergences \cite{nowozin2016f} or replaced by the Earth-Mover distance \cite{arjovsky2017wasserstein}. Despite the success, GANs are notoriously difficult to train\cite{kodali2018on,arjovsky2017towards}. When the support of these two distributions are approximately disjoint, the gradient given by the discriminator with the standard objective may vanish \cite{arjovsky2017wasserstein}. More seriously, the generated distribution can fail to cover the whole data distribution and collapse to a single mode in some cases \cite{dumoulin2017adversarially,che2016mode}.

\begin{figure}[H]
    \begin{center}
        \includegraphics[width=0.6\textwidth]{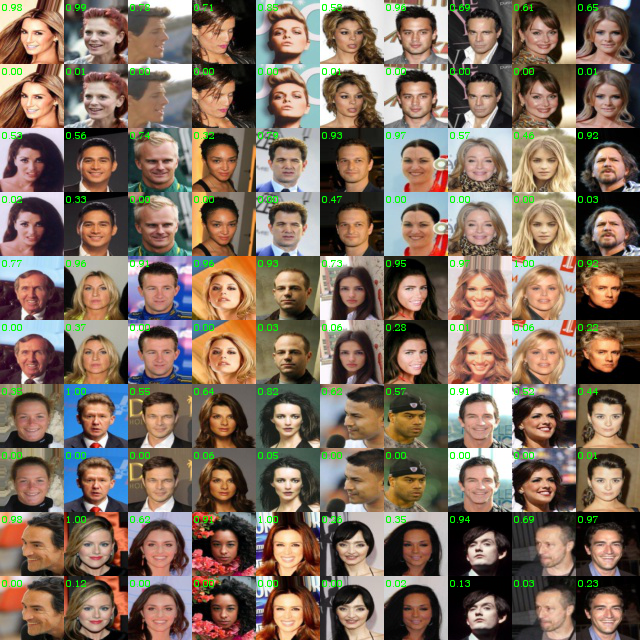}
    \end{center}
    \caption{Benign samples (on odd rows) and adversarial samples of the standard discriminator (on even rows). The confidence of the discriminator is depicted at the corner. Obviously, the standard discriminator is extremely vulnerable to imperceptible perturbation. The $L_{\infty}$ norm of the perturbation level is 1/255.}
    \label{adversarial_samples}
\end{figure}

\begin{figure}[h]

    \begin{center}
        \includegraphics[scale=0.6]{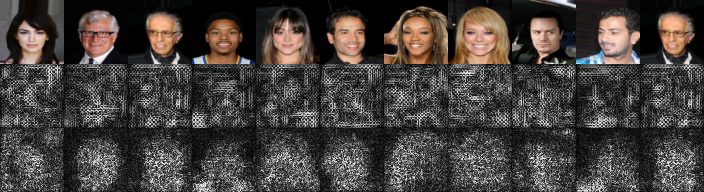}
    \end{center}
    \caption{Visualization of the gradient of the DCGAN discriminator with respect to input images. The first row shows samples from CelebA dataset. The second row and third row show the gradients of the standard discriminator and the adversarially trained discriminator, respectively. We clip the gradients to within $\pm$ 3 standard deviations of their mean and take the average absolute value of three channels for visualization.}
    \label{gradient_visual}
\end{figure}

From a practical standpoint, the gradient of the discriminator is a direct factor that affects parameter update of the generator,  which determines the training stability and performance to a great extent. However, the representation capacity of a discriminator realized by a neural network is not infinite. Meanwhile, the discriminator is usually not optimal to measure the true discrepancy when trained in an alternative manner. As a consequence, the discriminator is extremely vulnerable to adversarial samples \cite{szegedy2014intriguing,goodfellow2015explaining}, which can be validated by the experiments shown in Figure \ref{adversarial_samples}. Adding imperceptible perturbation to real samples can mislead the classifier to give wrong predictions. Adversarial samples can be easily crafted by gradient based methods such as Fast Gradient Sign Method (FGSM) \cite{goodfellow2015explaining} and Basic Iterative Method (BIM) \cite{kurakin2017adversarial}. It should be noted that the gradient given by the discriminator that guides update of the generator is exactly the same as the gradient used to craft adversarial samples of the discriminator. In other words, the gradient contains non-informative adversarial noise which is imperceptible but can mislead the generator and make training unstable.
Because the discriminator is adversarially trained with diverse generated fake samples, the misleading effect of adversarial noise is partly alleviated and vanilla GANs can still successfully generate meaningful samples instead of adversarial noise. However, the training is usually unstable. We find that this is partly because adversarial training on real samples, or training the discriminator with adversarial samples of real data, has been neglected, which results in the gradient of the discriminator still containing considerable adversarial noises. To tackle this issue, in this work, we propose adversarial symmetric GANs (AS-GANs) that perform adversarial training on both fake samples and real samples. Figure \ref{gradient_visual} visualizes and compares the gradient of the standard discriminator and the discriminator further adversarially trained on real samples. Obviously, the gradients of the standard discriminator appear to be non-informative noises,  while the gradients of the further adversarially trained discriminator contain more semantic information, such as profile of face, indicating that AS-GANs can effectively eliminate adversarial noise contained in the gradient of the discriminator.

We further verify AS-GANs on image generation with widely adopted DCGAN \cite{Radford2015Unsupervised} and ResNet \cite{he2015deep,gulrajani2017improved} architecture and obtained consistent improvement of training stability and acceleration of convergence. More importantly, FID scores of the generated samples are improved by $10\% \sim 50\%$ compared to the baseline on CIFAR-10, CIFAR-100, CelebA, and LSUN, demonstrating that the further adversarial trained discriminator provides more informative gradient with less adversarial noise. This work provides a new approach to further develop adversarial networks from the perspective of adversarial samples. Our main contributions are summarized as follows:
\begin{enumerate}[(1)]
    \item  We showed that the gradients of the discriminator contain non-informative noise and adversarial training on real samples has been overlooked in vanilla GANs from the perspective of adversarial samples.
    \item We proposed AS-GAN that incorporates adversarial training both on fake samples and real samples and provided both intuitive and theoretical justifications.
    \item We verified AS-GAN on image generation tasks with frequently-used networks on CIFAR-10, CIFAR-100, CelebA, and LSUN. AS-GAN with spectral normalization achieves the state-of-the-art performance on CelebA and LSUN.
\end{enumerate}

\section{Related Work}
A large body of work orthogonal to our method has been reported on how to stabilize GANs training by regularization and improving training strategy. Heusel et.al \cite{Heusel2017GANs} proposed to use individual learning rate for both the discriminator and the generator, which showed improved training but cannot solve training collapse completely. Kodali et.al \cite{kodali2018on} analysed the convergence of GAN training from the point of regret minimization and proposed to regularize the discriminator around real data samples with gradient penalty. Metz et.al \cite{metz2016unrolled} proposed to unroll the optimization of the discriminator as a surrogate objective to guide update of the generator, which showed improvement of training stability with a relatively large computation overhead.

Another line of work seeks to replace the Jensen-Shannon divergence used in vanilla GANs with integral probability metric.  Wasserstein GAN \cite{arjovsky2017wasserstein} and its variants \cite{gulrajani2017improved,wu2018wasserstein} can solve gradient vanishing in GANs training theoretically but the discriminator needs to meet 1-Lipschitz constraint.  Wasserstein GAN \cite{arjovsky2017wasserstein} adopts weight clip to make the discriminator 1-Lipschitz constrained but the capacity of the discriminator is significantly constrained. WGAN-GP \cite{gulrajani2017improved} adopts gradient penalty to regularize the discriminator in a less rigorous way, but it requires calculation of the second order derivative with  remarkable computation overhead. Spectral normalization on the weight of the discriminator proposed by \cite{Miyato2018Spectral} makes the discriminator 1-Lipschitz constrained efficiently but regularizes the network in a layer-wise manner.

Adversarial vulnerability is an intriguing property of neural network based classifiers \cite{szegedy2014intriguing}. A well-trained neural network can make completely wrong predictions on adversarial samples that human can recognize accurately. Small-magnitude adversarial perturbation added to benign data can be easily calculated based on the gradient \cite{goodfellow2015explaining,carlini2017towards,dong2018boosting}. Goodfellow et.al \cite{goodfellow2015explaining} proposed to augment training data with adversarial samples to improve the robustness of neural networks, which can smooth the decision boundary of classifiers around training samples. The gradients of adversarially trained classifiers contain more semantic information and less adversarial noise \cite{tsipras2018robustness,Kim2019Bridging}.

Some work tried to craft or defense against adversarial samples using GANs, which is different from our motivation. Xiao et.al \cite{Xiao2018Generating} proposed to generate adversarial samples efficiently with GANs, in which a generator is used to generate adversarial perturbation for the target classifier given original samples. Shen et.al \cite{Shen2017AE} proposed AE-GAN to eliminate adversarial perturbation in an adversarial training manner, generating images with better perceptual quality. Zhou et.al \cite{zhou2018dont} proposed to perform additional adversarial training on fake samples to robustify the discriminator.

RGAN \cite{zhang2020robust} is an independent work in which the generator and discriminator compete with each other in a worst-case setting within a small Wasserstein ball. Adversarial training on real samples and fake samples are all considered in RGAN. However, our work shows that adversarial training on fake samples has been considered in the standard setting. It is unnecessary to further train the discriminator explicitly with perturbed fake samples because the further improvement is marginal at the cost of additional computation overhead. Moreover,in RGAN, the generator is also trained with adversarial samples of noise. However, the vulnerability of the generator as a regression network is less serious than that of the discriminator. Therefore, in AS-GAN, we pay more attention on the part of the discriminator.

\section{Method}
In this section, we first review the training details of vanilla GANs. Then, from the perspective of adversarial samples, we identify that only adversarial training on fake samples is implemented in vanilla GANs, while that on real samples has been neglected. Thus, we introduce adversarial training of the discriminator on real samples into GANs, by which the adversarial training scheme becomes symmetrical and the standard objective of vanilla GANs is generalized to a robust one. As a result, adversarial noises contained in the gradient of the discriminator are further eliminated, facilitating more stable training. Additionally, we discuss the effect of perturbation level on performance intuitively.

\begin{figure}[h]

    \begin{center}
        \includegraphics[scale=0.6]{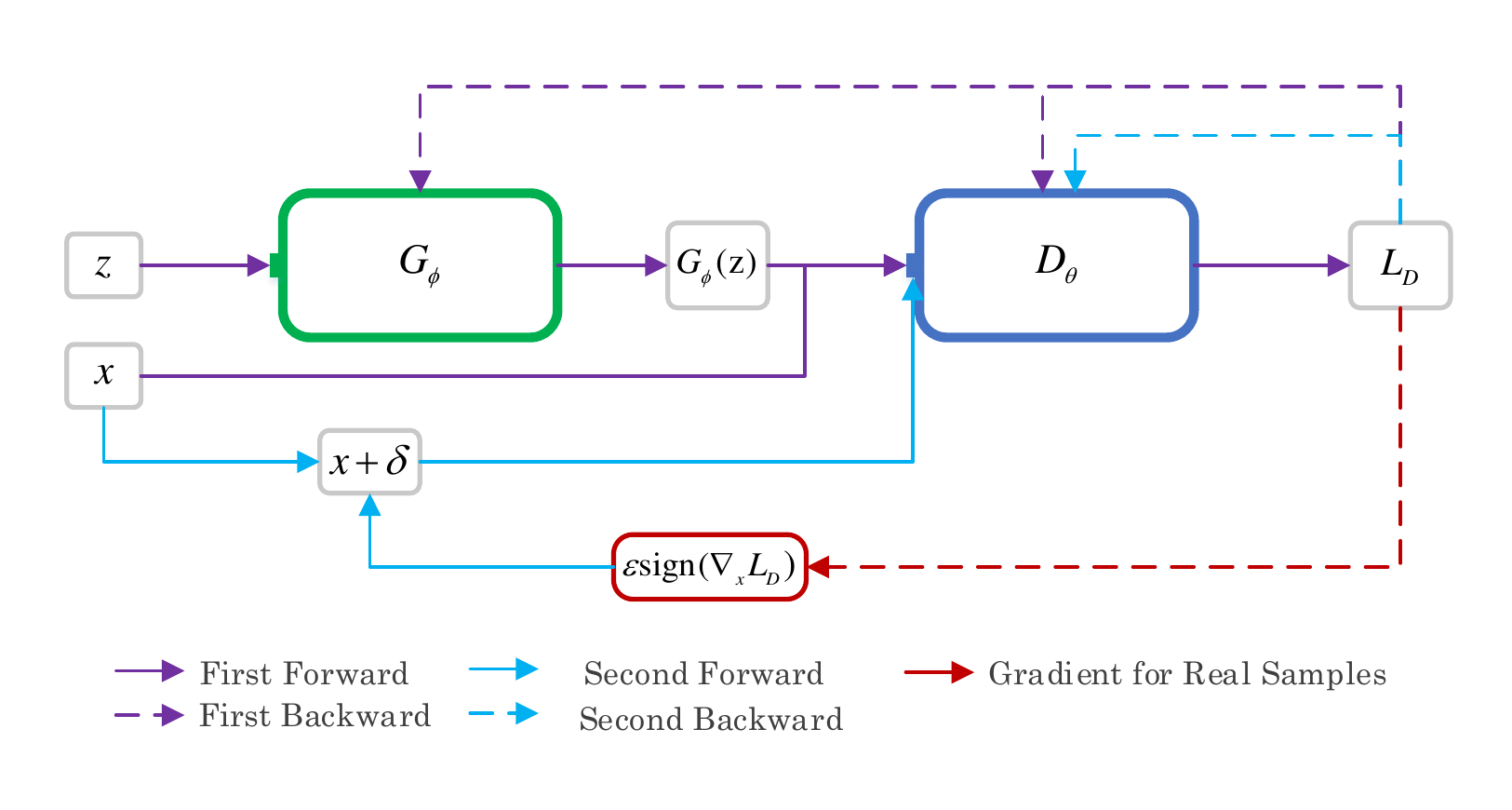}
    \end{center}
    \caption{An illustration of the proposed AS-GANs. The standard GAN training procedure is illustrated as the first forward pass and backward pass. In addition to that, we introduce adversarial training of the discriminator on real samples, illustrated as the second forward pass and backward pass, which is approximately equivalent to training the discriminator with robust optimization.}
    \label{Schematic}
\end{figure}
\subsection{Vanilla GAN}
In the GAN proposed by \cite{Goodfellow2014Generative}, the generator $G_{\phi}(z) $ parameterized by $\phi$ tries to generate fake samples to fool the discriminator and the discriminator $D_{\theta}(x)$ parameterized by $\theta$ tries to distinguish between generated samples and real samples. The formulation using min-max optimization is as follows:
\begin{equation}
    \label{min-max}
    \min_{\phi}\max_{\theta}V(\theta,\phi)
\end{equation}
where $V(\theta,\phi) $ is the objective function. Equation \ref{min-max} can be formulated as a binary classification problem with cross entropy loss:

\begin{equation}
    V(\theta,\phi) =\mathbb{E}_{x \sim P_{data}}\left[\log D_{\theta}(x)\right]+\mathbb{E}_{z\sim \mathcal{N}(0,I)} \left[ \log D_{\theta}(1-G_{\phi}(z))\right]
\end{equation}
where $P_{data}$ is the real data distribution and noise  $z$ obeys standard Gaussian distribution. When the discriminator is trained to be optimal, the training objective for the generator can be reformulated as the Jensen-Shannon divergence, which can measure dissimilarity between two distributions.

In practice, we use mini-batch gradient descent to optimize the generator and the discriminator alternatively. At each iteration, update rule can be derived as follows:
\begin{equation}
    \label{D_update}
    \begin{split}
        \theta'&=\theta+\eta_{\theta} \nabla_{\theta}V_{m}(\theta,\phi,x,z)\\
        \phi'&=\phi-\eta_{\phi}\nabla_{\phi}V_{m}(\theta,\phi,x,z)
    \end{split}
\end{equation}
where $\eta_{\theta}$ and $ \eta_{\phi}$ are the learning rate of the discriminator and the generator, respectively. $V_{m}(\theta,\phi,x,z) $ denotes the objective function of mini-batch with $m$ real samples and $m$ fake samples, which is:
\begin{equation}
    \label{mini-batch objective}
    V_{m}(\theta,\phi,x,z)=\frac{1}{m}\sum_{i=1}^{m}\left[ \log D_{\theta}(x^i)+\log D_{\theta}(1-G_{\phi}(z^i))\right]
\end{equation}
After the parameters of networks are updated, fake samples generated by $G_{\phi'}(z)$ are adjusted as following equation according to the chain rule:
\begin{equation}
    \begin{split}
        G_{\phi'}(z) &\approx G_{\phi}(z)-\eta_{\phi}\frac{\partial G_{\phi}(z) }{\partial \phi}\nabla_{\phi}V_{m}(\theta,\phi,x,z)                                                    \\
        & =G_{\phi}(z)-\eta_{\phi}\frac{\partial G_{\phi}(z) }{\partial \phi}(\frac{\partial G_{\phi}(z) }{\partial \phi})^T\nabla_{G_{\phi}(z)}V_{m}(\theta,\phi,x,z)
    \end{split}
\end{equation}
where $\frac{\partial G_{\phi}(z) }{\partial \phi}$ is a Jacobian matrix. The updated $G_{\phi'}(z)$ can be seen as adversarial samples of the discriminator at this iteration because $\eta_{\phi}$ is usually small. These samples will be fed into the discriminator at future iterations to perform adversarial training. From this point of view, we find that vanilla GANs mainly include adversarial training on fake samples, which is illustrated as the first pass in Figure \ref{Schematic}. However, this framework ignores adversarial training of discriminator on real samples, which makes the gradient of the discriminator contain considerable adversarial noise because of the unsmooth decision boundary of the discriminator.

\subsection{Adversarial training on real samples}
To robustify the discriminator and reduce adversarial noise, we incorporate adversarial training on real samples into vanilla GANs. Specifically, we perform adversarial training after Equation \ref{D_update} at each iteration as the following equation:
\begin{equation}
    \theta'=\theta+\eta_{\theta}\nabla_{\theta}V_{m}(\theta,\phi,\hat{x},z)
\end{equation}
where $\hat{x}$ is an adversarial sample of the discriminator, the perturbation of which is the gradient of the discriminator with respect to $x$. The adversarial sample can be calculated with constant $\varepsilon$ as follows:
\begin{equation}
    \hat{x}=x-\varepsilon \sign(\nabla_{x}V_{m}(\theta,\phi,x,z))
\end{equation}
This adversarial training formulation is adopted from \cite{goodfellow2015explaining}, which calculates $L_{\infty}$-norm constrained perturbation by linearizing the objective function.

Adversarial training on real samples of the discriminator is illustrated as the second pass in Figure \ref{Schematic}, where $L_{D}$ denotes the minus of $V_{m}(\theta,\phi,x,z)$. The reason of adopting this training scheme is two fold: First, the gradient used to craft perturbation can be obtained from the first backward pass conveniently and the overall computation overhead of additional training is relatively low, about 25\%. Second, despite simplicity, this scheme provides significant improvement. We have also conducted experiments on other more complicated adversarial training schemes such as Projected Gradient Descent (PGD) attack \cite{MadryTowards}, which achieve comparable improvement to that of FGSM. Actually, we can feed both real samples and adversarial samples at one pass, but this requires an additional pass to calculate adversarial perturbation, which is computational-inefficient. Therefore, we train the discriminator with real samples and adversarial samples in separate passes, respectively.

Please refer to Algorithm \ref{algorithm_table} for more details about symmetric adversarial training.

\begin{algorithm}[H]
    \caption{Mini-batch stochastic gradient descent training of AS-GAN. We set perturbation $\varepsilon$ to $1/255$ as default for image generation tasks.}
    \begin{algorithmic}[1]
        \label{algorithm_table}
        \FOR{number of training iterations}
        \STATE Sample mini-batch of $m$ noise samples $\{z^i,...,z^m\}$ from Gaussian distribution $\mathcal{N}(0,I)$.\
        \STATE Sample mini-batch of $m$ data samples $\{x^i,...,x^m\}$ from real data distribution $P_{data}$.\
        \STATE Update discriminator by gradient ascent:\
        \begin{equation}
            \theta'=\theta+\eta_{\theta}\nabla_{\theta}V_{m}(\theta,\phi,x,z)
        \end{equation}
        \STATE \textbf{Craft adversarial samples of real samples for discriminator}:
        \begin{equation}
            \hat{x}=x-\varepsilon \sign(\nabla_{x}V_{m}(\theta,\phi,x,z))
        \end{equation}
        \STATE \textbf{Perform adversarial training of discriminator on real samples}:
        \begin{equation}
            \theta'=\theta+\eta_{\theta}\nabla_{\theta}V_{m}(\theta,\phi,\hat{x},z)
        \end{equation}
        \STATE Update generator by gradient descent:
        \begin{equation}
            \phi'=\phi-\eta_{\phi}\nabla_{\phi}V_{m}(\theta,\phi,x,z)
        \end{equation}
        \ENDFOR
    \end{algorithmic}
\end{algorithm}

\newpage

\subsection{Effective perturbation level}
The perturbation level is a crucial hyper parameter that affects the effectiveness of adversarial training. When perturbation is set to zero, our AS-GANs degrade to updating discriminator twice on the same real data. When the perturbation level is set too large, the real data will be drastically perturbed. Semantic information and quality will be changed, which misleads the discriminator to recognize degraded samples as real. In addition, we recommend setting the perturbation to zero at the beginning of training in case that the discriminator is too weak. The next section will cover extensive experiments on how perturbation affects training.

\subsection{Adversarial training on fake samples}
In standard GAN training, the generator is updated to generate adversarial samples of fake samples for the discriminator in each iteration. In the next iteration, the generator might not generate examples as aggressive as adversarial example for the current discriminator, because the generator is trained with the last batch of $z$ and the new batch of $z$ is changed. This seems to indicate that the adversarial training of the discriminator on fake samples is not very strong. However, from an overall consideration of performance improvement and computation cost, we are convinced that it is not necessary to further train the discriminator with adversarial samples of fake samples in each iteration. The reasons are as follows.

It was reported that adversarial vulnerability partially attributes to the limited training data \cite{2018Adversarially}. For the discriminator, training data of fake samples generated by the generator with the input of random noise is indeed infinite, but the training data of real samples is limited. This makes the discriminator less vulnerable to adversarial samples of fake samples although the adversarial training of the discriminator on fake samples is not very strong. Training the discriminator with adversarial samples of fake samples may further improve the robustness of the discriminator. However, from a practical point of view, the cost performance is low. This can be verified by the experimental results shown in Table \ref{adv_samples_comparison}. The FID of AS-DCGAN-rf that is adversarially trained on both real and fake samples is 1.1\% higher than that of AS-DCGAN-r that is adversarially trained on real samples, but the training time of one epoch increases by 15.8\%. Hence, if the training cost is not a critical issue, it can certainly incorporate adversarial training of the discriminator on fake samples into training GANs, although the additional improvement is marginal.

\begin{table}[h]
    \centering
    \begin{threeparttable}
        \caption{Experimental results of DCGAN trained with different adversarial samples}
        \label{adv_samples_comparison}
        \begin{tabular}{cccc}
            \toprule
            Setting                 & DCGAN   & AS-DCGAN-r  & AS-DCGAN-rf  \\
            \midrule
            FID                     & 28.60  & 26.54  & 26.24     \\
            Training time per epoch & 19.83s & 26.40s & 30.58s  \\
            \bottomrule
        \end{tabular}
    \end{threeparttable}
\end{table}

\section{Experiments}
To evaluate our method and identify the reasons behind its efficacy, we test our adversarial training method on image generation on CIFAR-10, CIFAR-100, CelebA, and LSUN with DCGAN and ResNet architecture. CIFAR-10 is a well-studied dataset of $32\times 32 $ natural images, containing 10 categories of objects and 60k samples. CIFAR-100 is the same as CIFAR-10 except for containing 100 categories. CelebA is a face attributes dataset with more than 200k images. LSUN is a large scale scene understanding dataset with 10 categories and we choose 3000k images labeled as bedroom for training. For fast validation, we resize images in CelebA and LSUN to $64\times 64 $. Before fed into discriminator, images are rescaled within $[-1,1]$. The dimension of the latent vector is set to 100 for all implementations.

In ResNet architecture, the residual block is organized as BatchNorm-ReLU-Resample-Conv-BatchNorm-ReLU-Conv with skip connection. We use bilinear interpolation for upsampling and average pooling for downsampling. Batch normalization \cite{Ioffe2015Batch} is used for both generator and discriminator. Parameters of network are initialized by Xavier method \cite{glorot2010understanding}.

In DCGAN architecture, the basic block is organized as Conv-BatchNorm-LeakyReLU for discriminator and ConvTransposed-BatchNorm-ReLU for generator. The weights of convolution are initialized by normal distribution with zero mean and 0.02 standard deviation. Bias is not used for convolution.

We implement models by Pytorch with acceleration of RTX 2080ti GPUs. Networks are trained with Adam optimizer \cite{kingma2014adam} with learning rate 2e-4. $\beta_1$ is set to 0.5 and $\beta_2$ is set to 0.999. Because training standard GANs on CelebA is unstable, we decrease the learning rate of the discriminator to 5e-5 to balance training as the TTUR training strategy \cite{Heusel2017GANs}.  We train models on CelebA for 100 epochs, CIFAR-10 for 200 epochs, and LSUN for 10 epochs.

In this paper, we use Fr\'echet inception distance (FID) \cite{Heusel2017GANs} and inception score \cite{Salimans2016Improved} to measure model performance, both of which are well-studied metric of image modeling. Our source code is available on Github \url{https://github.com/CirclePassFilter/AS_GAN}.

\subsection{Evaluation with different perturbation level}
\label{hyper-parameter}
In this section, we will investigate how perturbation level affects performance improvement. Specifically, we perform unsupervised image generation on CIFAR-10 with DCGAN architecture in different settings of perturbation level. Due to the large searching space, we select several typical values for experiments such as \{0,1,2,3,4\}/255. All experiments are conducted three times independently to reduce the effect of randomness.

\begin{figure}[H]
    \centering
    \subfigure[]{
        \includegraphics[width=0.45\textwidth]{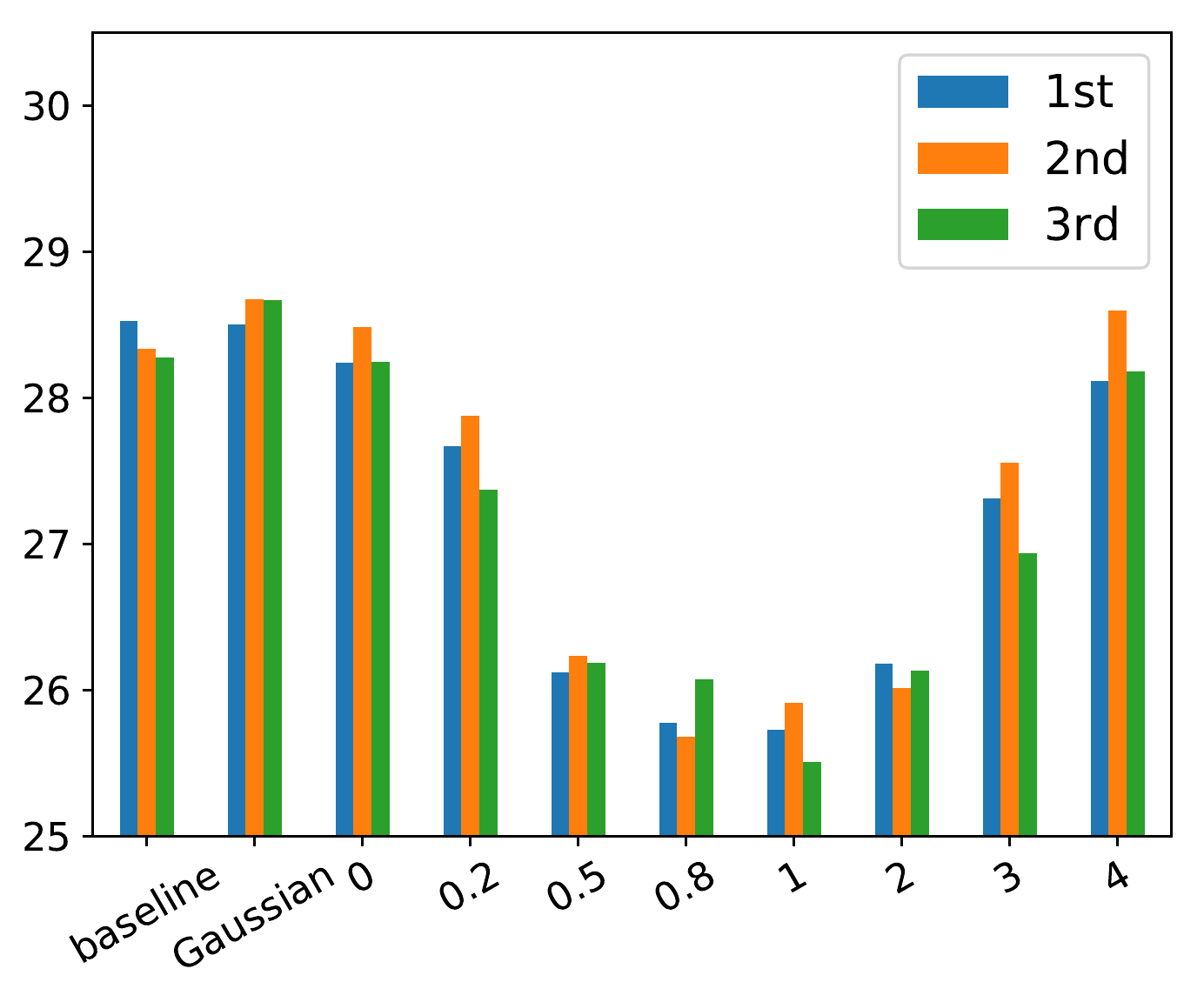}
    }
    \subfigure[]{
        \includegraphics[width=0.45\textwidth]{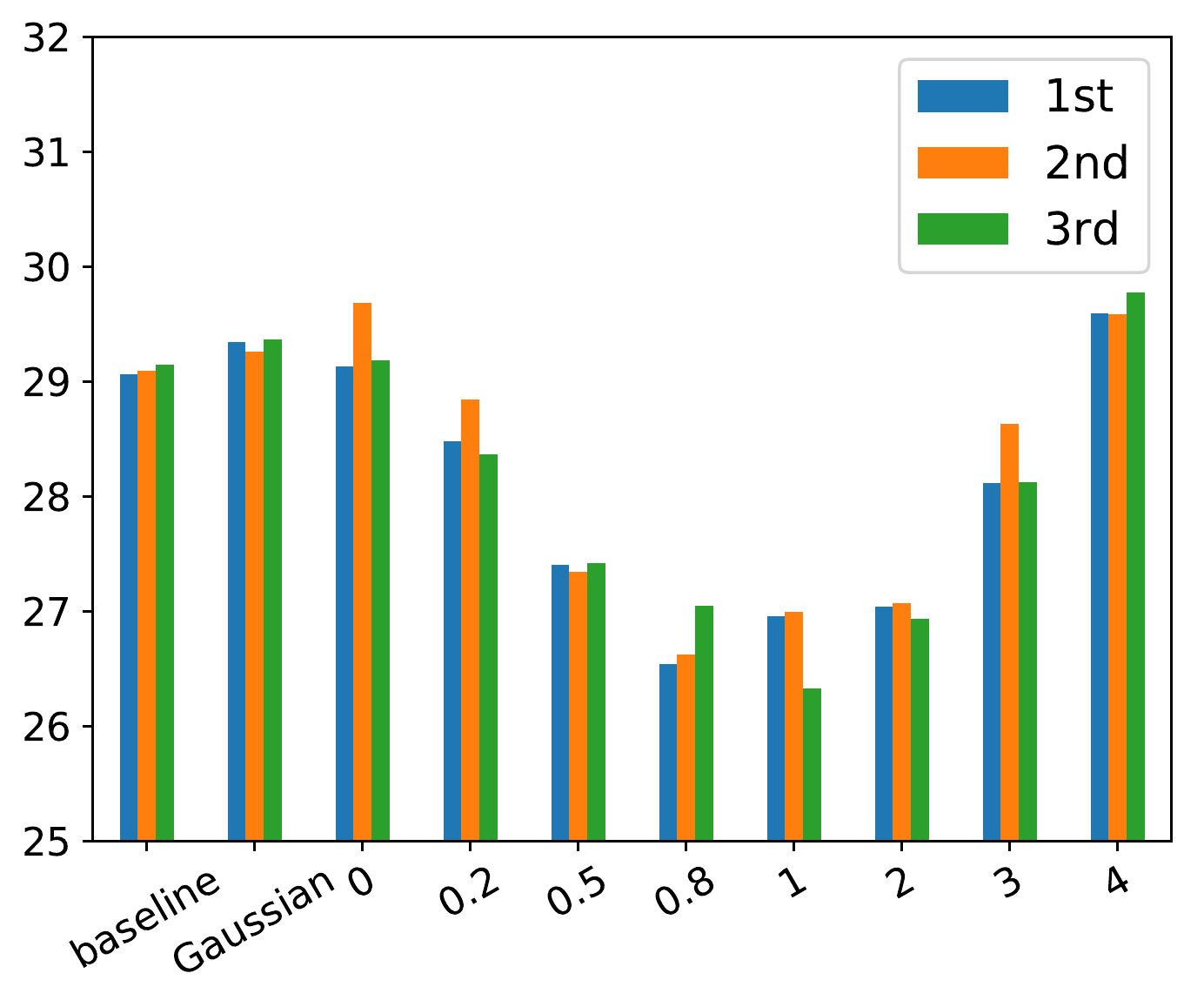}
    }
    \caption{(a) Best FID scores in different settings of three independent runs (Lower is better). (b) Mean FID scores of last 20 epochs in different settings of three independent runs. `Gaussian' means that the gradient used to craft adversarial samples is replaced by Gaussian noise.}
    \label{bar-fid}
\end{figure}

As shown in Figure \ref{bar-fid}, our method performs better than the baseline when $\varepsilon$ lies in interval $0.5/255\sim 3/255$. FID scores are improved by a large margin with perturbation of $1/255$. However, when the perturbation level is too tiny ($\varepsilon=0.2/255$), the method improves the original model marginally. In this case, the effect of adversarial training is limited. When the imposed perturbation is too strong, the model performs even worse than the baseline. This is because the discriminator is forced to recognize degraded samples as real and cannot provide the correct gradient to update the generator.

In short, with appropriate perturbation, the discriminator can be regularized to be more robust using AS-GANs, facilitating the generation of more accurate and informative gradient. Thus, the generator can obtain more reliable gradients, thereby reducing collapse of training and improving fidelity of generated samples.

\subsection{Evaluation with different adversarial manipulation methods}
In addition to FGSM \cite{goodfellow2015explaining}, various adversarial manipulation methods such as BIM \cite{kurakin2017adversarial} and PGD \cite{MadryTowards} have been proposed. Generally, these methods can be used to generate adversarial samples in AS-GAN and are expected to achieve more appealing improvement. We experiment AS-GAN with PGD at the different number of iterations on CIFAR-10. For comparison, other experiment settings such as network architecture and hyper parameters remain the same as that of AS-GAN with FGSM. The maximum perturbation is set to 1/255. When the number of iterations is set to 1, PGD method degrades to FGSM with random initialization. The experimental results are shown in Table \ref{pgd_comparison}. AS-GAN with PGD can achieve more significant improvement compared to that with FGSM although the training time increases linearly with the number of iterations. In the following experiments, we use FGSM as the default adversarial manipulation method for the sake of efficiency.

\begin{table}[h]
    \centering
    \begin{threeparttable}
        \caption{Experimental results of AS-GAN with PGD}
        \label{pgd_comparison}
        \begin{tabular}{ccccccc}
            \toprule
            Setting                 & FGSM   & PGD-2  & PGD-4  & PGD-6  & PGD-12 \\
            \midrule
            FID                     & 25.50  & 25.31  & 24.99  & 25.09  & 25.12  \\
            Training time per epoch & 26.40s & 34.09s & 42.10s & 54.63s & 80.41s \\
            \bottomrule
        \end{tabular}
    \end{threeparttable}
\end{table}

\subsection{Evaluation on variants of Wasserstein GANs }
The proposed method of the vanilla GAN is easily extended to Wasserstein distance based GANs. We conduct experiments of the extension of AS-GAN on WGAN-GP, named as AS-WGAN-GP, on CIFAR-10 with DCGAN architecture. The experimental results in the following Table \ref{wgangp_as} show that AS-WGAN-GP performs better than WGAN-GP although the improvement is marginal. This is because the proposed method has a similar effect on constraining the Lipschitz constant of the discriminator. On the other hand, the discriminator of WGAN used to measure the Wasserstein distance is a regression network, whose vulnerability is less serious than classification networks. Therefore, the proposed method is more suitable for the GAN whose discriminator is a classifier. AS-GAN with vanilla objective can achieve the state-of-the-art performance on some benchmark datasets.

\begin{table}[h]
    \centering
    \begin{threeparttable}
        \caption{Experimental results of AS-WGAN-GP}
        \label{wgangp_as}
        \begin{tabular}{ccc}
            \toprule
            Model & WGAN-GP & AS-WGAN-GP \\
            \midrule
            FID   & 37.22   & 36.62      \\
            \bottomrule
        \end{tabular}
    \end{threeparttable}
\end{table}

\subsection{Ablation study}
\label{ablation-study}
\begin{figure}[H]
    \centering
    \includegraphics[width=0.48\textwidth]{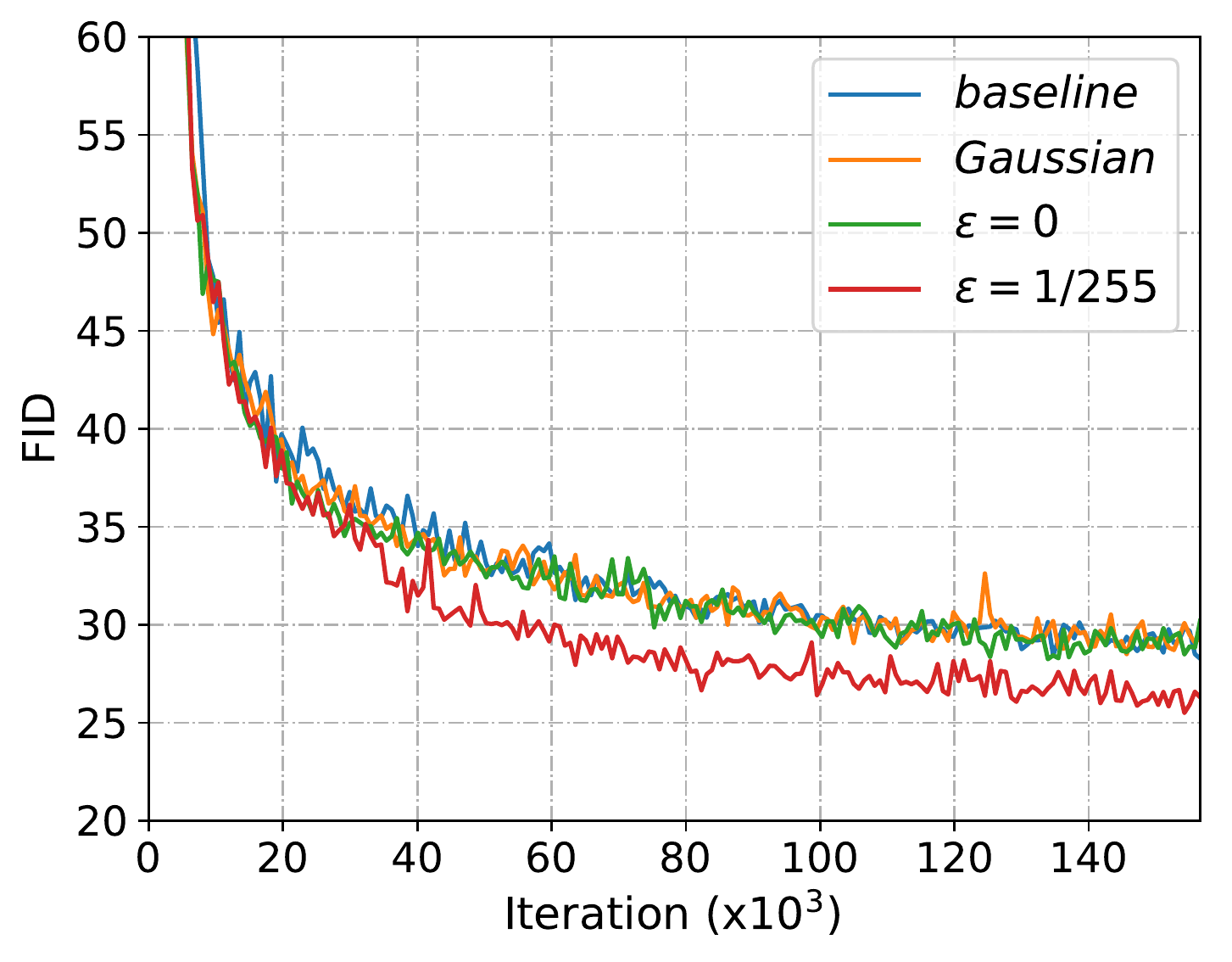}
    \caption{Training curves of FID in different settings. Updating the discriminator twice on the same data ($\varepsilon=0$) or perturbing samples with random noise cannot work, which indicates that the comparison between the proposed method and the baseline is fair.}
    \label{conver}
\end{figure}

In addition, we plot the training curve of FID on CIFAR-10 as shown in Figure \ref{conver}. In particular, when $\varepsilon$ is zero, AS-GANs degrade to updating the discriminator twice on the same real data, and the results are similar to the baseline but much worse than that of the setting with appropriate perturbation. This indicates that the performance improvement achieved by AS-GANs does not attribute to additional update of the discriminator. Furthermore, we conduct another experiment, replacing the gradient used to craft perturbation with Gaussian noise. The FID score of this setting is slightly worse than the baseline, indicating that perturbation of gradient direction rather than random direction is a key factor that makes AS-GANs effective.

\subsection{Evaluation with different architectures}
\label{ARCHITECTURE}
\begin{figure}[H]
    \begin{center}
        \includegraphics[height=8.0cm,width=10.0cm]{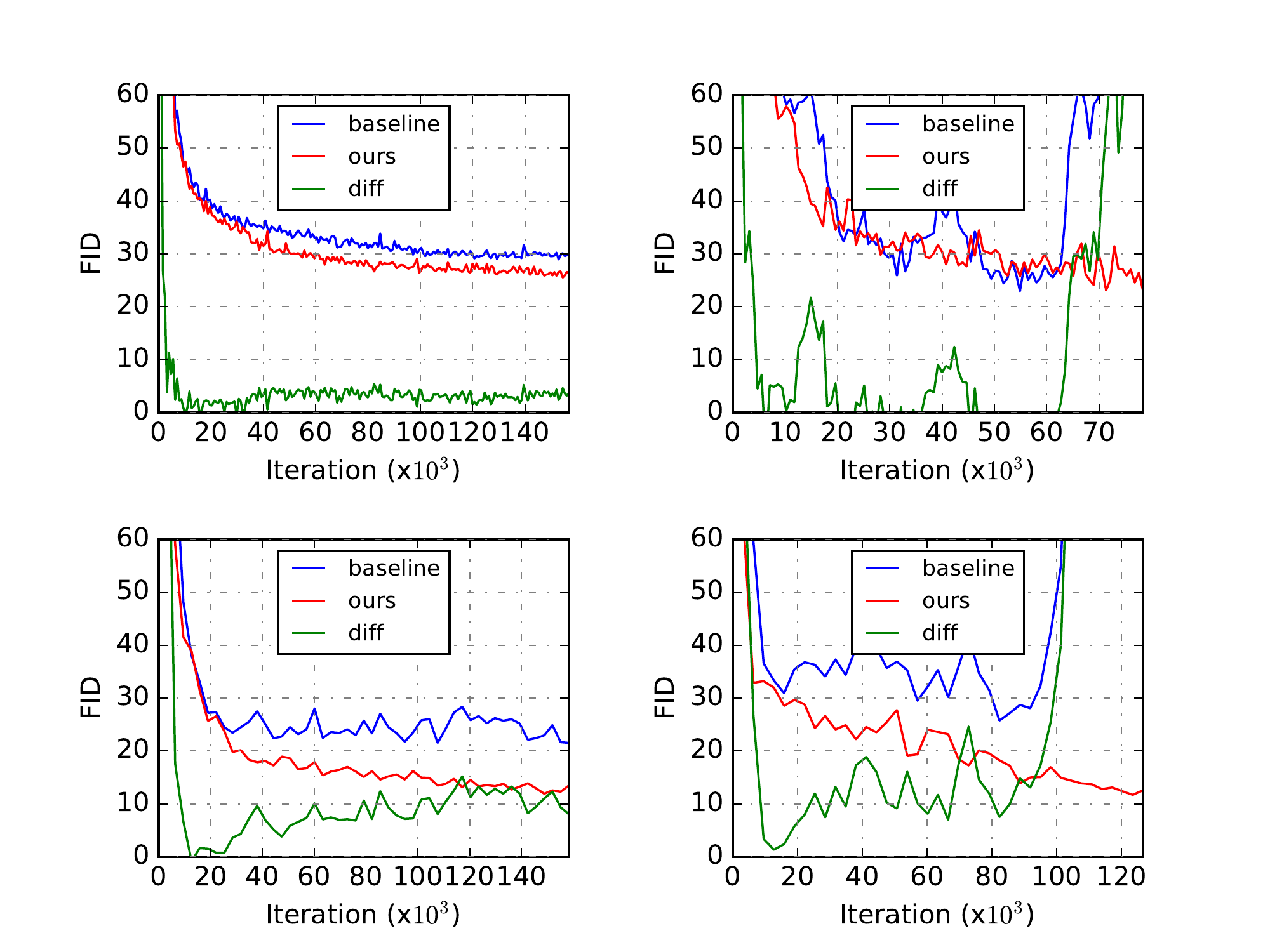}
    \end{center}
    \caption{Training curves of FID on CIFAR-10 (upper) and CelebA (lower) with DCGAN (left) and ResNet (right). The FID difference between the AS-GAN and the baseline is plotted as the green curve. Results show that AS-GANs can accelerate convergence and achieve better FID scores. Meanwhile, it can stabilize training with less sensitivity to network architecture and hyper parameter setting.
    }
    \label{architecture}
\end{figure}

To explore the compatibility of AS-GANs, we test it with widely adopted DCGAN and ResNet architecture on CIFAR-10 and CelebA. The comparison results plotted in Figure \ref{architecture} indicate that, with AS-GANs, training is more stable and the convergence is accelerated. Even at the setting that causes vanilla GANs collapse, our model still converges stably. The FID scores of the proposed method are significantly improved by $50\%$ on CelebA and $30\%$ on LSUN.

More importantly, we further combined AS-GANs with spectral normalization and achieved a FID of 5.88 on CelebA, which exceeds the FID of 11.71 obtained by using AS-GANs alone. This indicates that our AS-GANs can be combined with other schemes to further improve performance in practical applications.

\begin{figure}[H]
    \centering
    \subfigure[]{
        \includegraphics[width=0.4\textwidth]{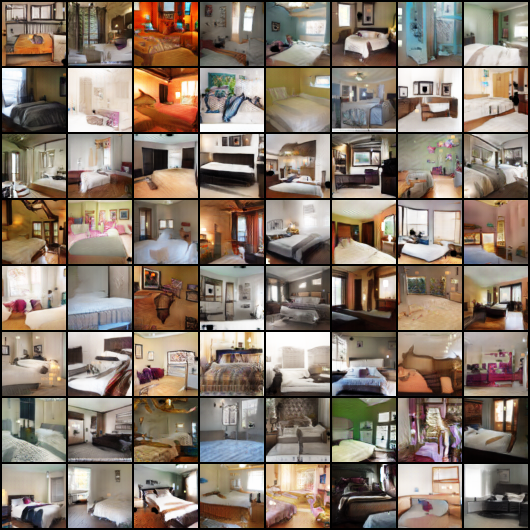}
    }
    \subfigure[]{
        \includegraphics[width=0.4\textwidth]{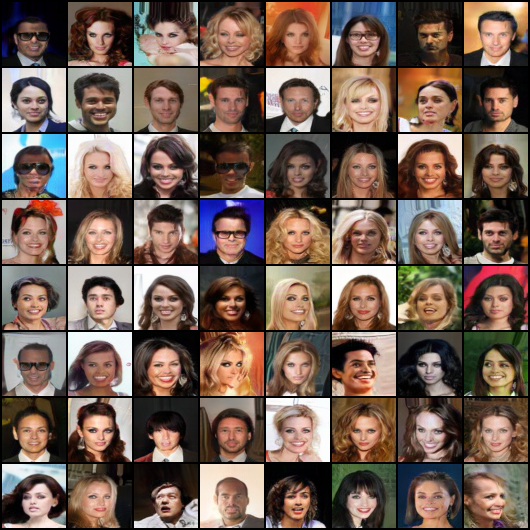}
    }
    \caption{(a): $64\times 64$ LSUN samples generated by AS-DCGAN. (b): $64\times 64 $ CelebA samples generated by AS-ResNet. }
    \label{generated_samples_app}
\end{figure}

\begin{figure}[H]
    \centering
    \subfigure[]{
        \includegraphics[width=0.4\textwidth]{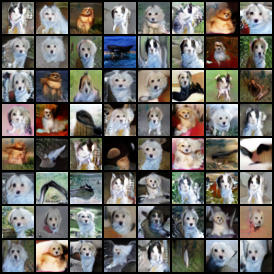}
    }
    \subfigure[]{
        \includegraphics[width=0.4\textwidth]{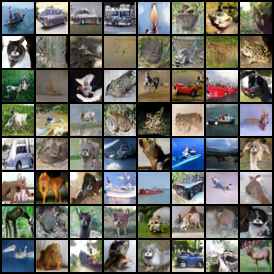}
    }
    \caption{(a): Collapsed samples generated by standard GAN trained on CIFAR-10. (b): Samples generated by AS-ResNet trained on CIFAR-10.}
    \label{mode}
\end{figure}

The outcomes of the unsupervised training on CIFAR-10, CIFAR-100, CelebA and bedroom in LSUN are summarized in Table \ref{table_performance}. The six rows at the bottom are the results of our implementation. It clearly demonstrates that, the AS-GANs with spectral normalization achieve comparable performance to the state-of-the-art. The generated samples are shown in Figure \ref{generated_samples_app}.  What's more, we observe that the proposed method can alleviate mode collapse to a large extent as illustrated in Figure \ref{mode}.

\renewcommand{\arraystretch}{0.8}
\begin{table}[H]
    \centering
    \begin{threeparttable}
        \caption{Inception scores and FIDs of unsupervised image generation. $\dag $\cite{Miyato2018Spectral}, $\ddag $  \cite{Wu2017Wasserstein}, $\ast $\cite{gulrajani2017improved} }
        \label{table_performance}
        \begin{tabular}{cccccc}
            \toprule
            \multirow{2}{*}{Method}             &
            \multicolumn{1}{c}{Inception score} & \multicolumn{4}{c}{FID}\cr
                                                & CIFAR-10                   & CIFAR-10      & CIFAR-100 & CelebA        & LSUN \cr
            \midrule
            (Standard CNN) \cr
            WGAN-GP                             & 6.68$\pm$.06$^\dag $       & 40.2$^\dag $  &           & 21.2$^\ddag $ & \cr
            SN-GAN$^\dag $                      & 7.58$\pm$.12               & 25.5          &           &               & \cr
            WGAN-GP(ResNet)                     & 7.86$\pm$.07$^\ast $       & 18.8$^\ddag $ &           & 18.4$^\ddag $ & 26.8$^\ddag $ \cr
            WGAN-div(ResNet)$^\ddag $           &                            & 18.1          &           & 15.2          & 15.9\cr
            \midrule
            DCGAN                               & 7.05$\pm$.14               & 28.05         & 28.60     & 20.45         & 25.36\cr
            AS-DCGAN                            & {\bf7.21}$\pm$.02          & {\bf25.50}    & \bf26.54  & {\bf 10.90}   & {\bf 18.08}\cr
            AS-SN-DCGAN                         & {\bf7.24}$\pm$.14          & {\bf24.50}    & \bf25.40  & {\bf 10.60}   & {\bf 21.84}\cr
            ResNet                              & 7.35$\pm$.16               & 22.92         & 25.63     & 25.72         & 175.70 \cr
            AS-GAN(ResNet)                      & {\bf7.65}$\pm$.15          & {\bf21.84}    & \bf25.38  & {\bf11.71}    & {\bf 45.96}\cr
            AS-SN-GAN(ResNet)                   & {\bf7.84}$\pm$.17          & {\bf22.26}    & \bf23.60  & {\bf 5.88}    & {\bf 8.00}\cr
            \bottomrule
        \end{tabular}
    \end{threeparttable}
\end{table}

\section{Analysis}

\subsection{Theoretical analysis}
By introducing adversarial training on real samples, we generalize the original objective of GAN to the following one that forces the discriminator to be robust:
\begin{equation}
    \hat{V}(\theta,\phi) =\mathbb{E}_{x\sim P_{data}}\left[\min_{\Vert \delta \Vert_p \le \varepsilon}\log D_{\theta}(x-\delta)\right]+\mathbb{E}_{z \sim \mathcal{N}(0,I)} \left[ \log D_{\theta}(1-G_{\phi}(z))\right]
\end{equation}

Considering $\varepsilon$ is small, we get:
\begin{equation}
    \min_{\Vert \delta \Vert_p \le \varepsilon}\log D_{\theta}(x-\delta)\
    \approx \log D_{\theta}(x)-\min_{\Vert \delta \Vert_p \le \varepsilon}\left[\delta^T\nabla_{x}\log D_{\theta}(x)\right]
\end{equation}

By Lagrange multiplier method, the optimal solution of the above equation can be calculated as follows.
\begin{equation}
    \delta=\varepsilon\left(\frac{\nabla_{x}\log D_{\theta}(x)}
    {\Vert\nabla_{x}\log D_{\theta}(x)\Vert_{\frac{p}{p-1}}}\right)^{\frac{1}{p-1}}
\end{equation}

Hence, the generalized objective is composed of the original objective and a regularizer that forces the norm of $\nabla_{x}\log D_{\theta}(x)$ to be small. The relation between gradient regularization and adversarial training is also analysed in \cite{7373334,9133291}
\begin{equation}
    \hat{V}(\theta,\phi) \approx V(\theta,\phi)-
    \varepsilon\Vert\nabla_{x}\log D_{\theta}(x)\Vert_{\frac{p}{p-1}}
\end{equation}

Here $\varepsilon$ is the coefficient that balances the regularization and the original objective. In a sense, the proposed adversarial training method is an alternative that constrains the Lipschitz constant of the discriminator on real data, which can usually stablize GAN training and improve the performance \cite{Miyato2018Spectral,2017Loss,arjovsky2017wasserstein,kodali2018on}. As discussed in Lipschitz GANs \cite{Zhou2019Lipschitz}, the Lipschitz condition is a vital element to alleviate the gradient uninformativeness problem that the optimal discriminator usually tells nothing about the real distribution. Penalizing the Lipschitz constant makes the gradient informative, thus leading to stable convergence and improved performance.

Different from spectral normalization \cite{Miyato2018Spectral} that normalizes the weight matrices of the discriminator in a greedy manner, AS-GAN constrains the discriminator more gracefully. The gradient penalty term used in WGAN-GP is 1-centered and regularizes the discriminator on the linear interpolation of real data and fake data. On the contrary, the equivalent gradient penalty term of AS-GAN is 0-centered and regularizes the discriminator on real data. As discussed in [2], 0-centered gradient penalty on real data distribution leads to local convergence but 1-centered gradient penalty used in WGAN-GP fails, when trained on a prototypical dataset. This indicates that 0-centered gradient penalty may be more appropriate for stabilizing GAN training in general cases where the data and generator distributions lie on low dimensional manifolds.

In fact, when training data of real samples is infinite and the distribution of real data is continuous, the above objective is approximately equivalent to the original, which may be the possible reason why adversarial training of the discriminator on real samples has long been ignored by the community as it focuses more on ideal formulation of GANs.  However, in practice, contrary to the fake data, the number of real samples is always limited, which partly accounts for the existence of adversarial samples. In a sense, adversarial training on real samples regularizes the discriminator by augmenting training data, which can smooth the decision boundary of the discriminator and make the discriminator more robust. A robust discriminator can provide more informative gradient and  stablize training.

\subsection{Gradient visualization}
\label{gradient}
\begin{figure}[H]
    \centering
    \subfigure[]{
        \includegraphics[width=0.45\textwidth]{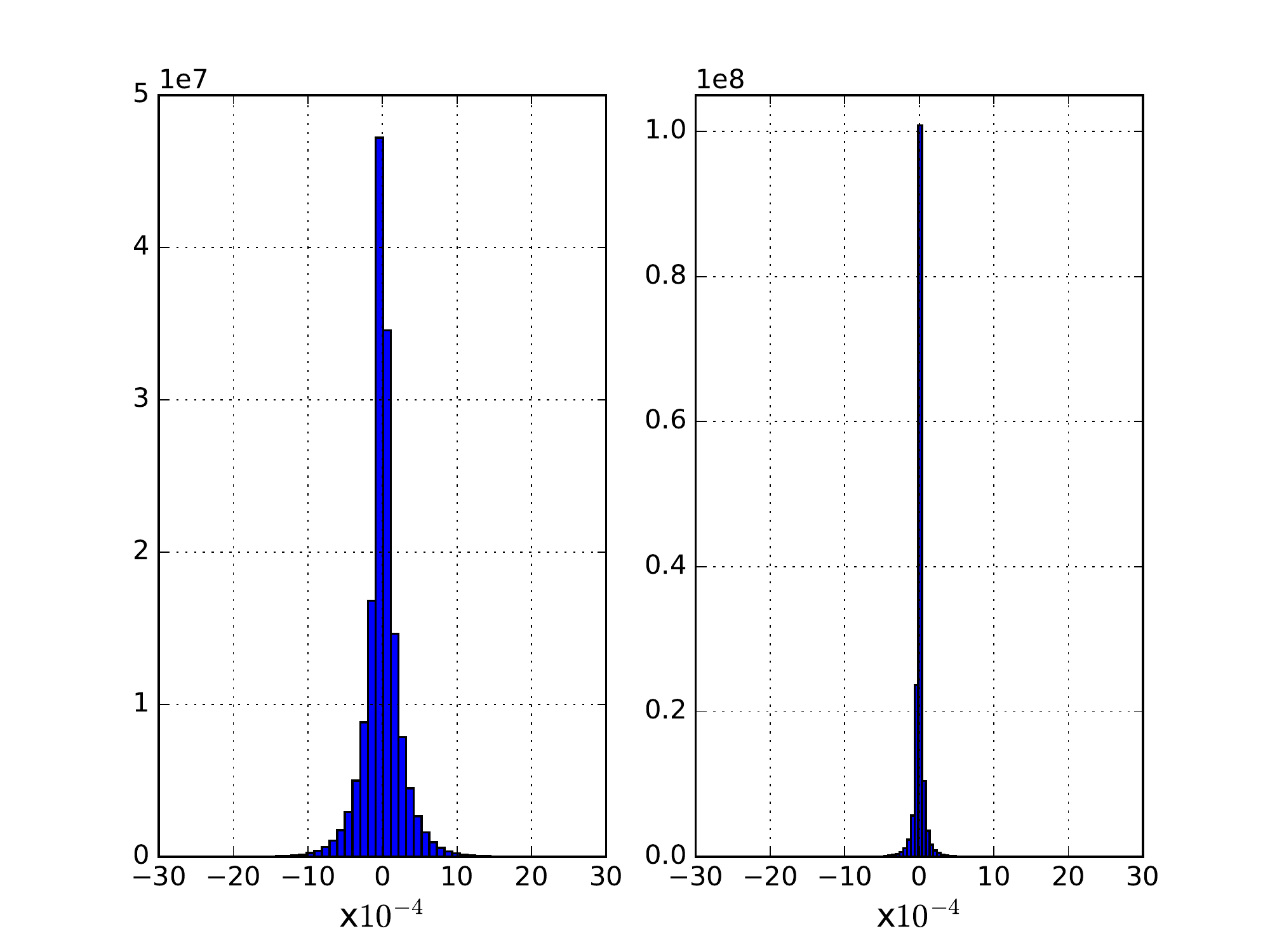}
    }
    \subfigure[]{
        \includegraphics[width=0.45\textwidth]{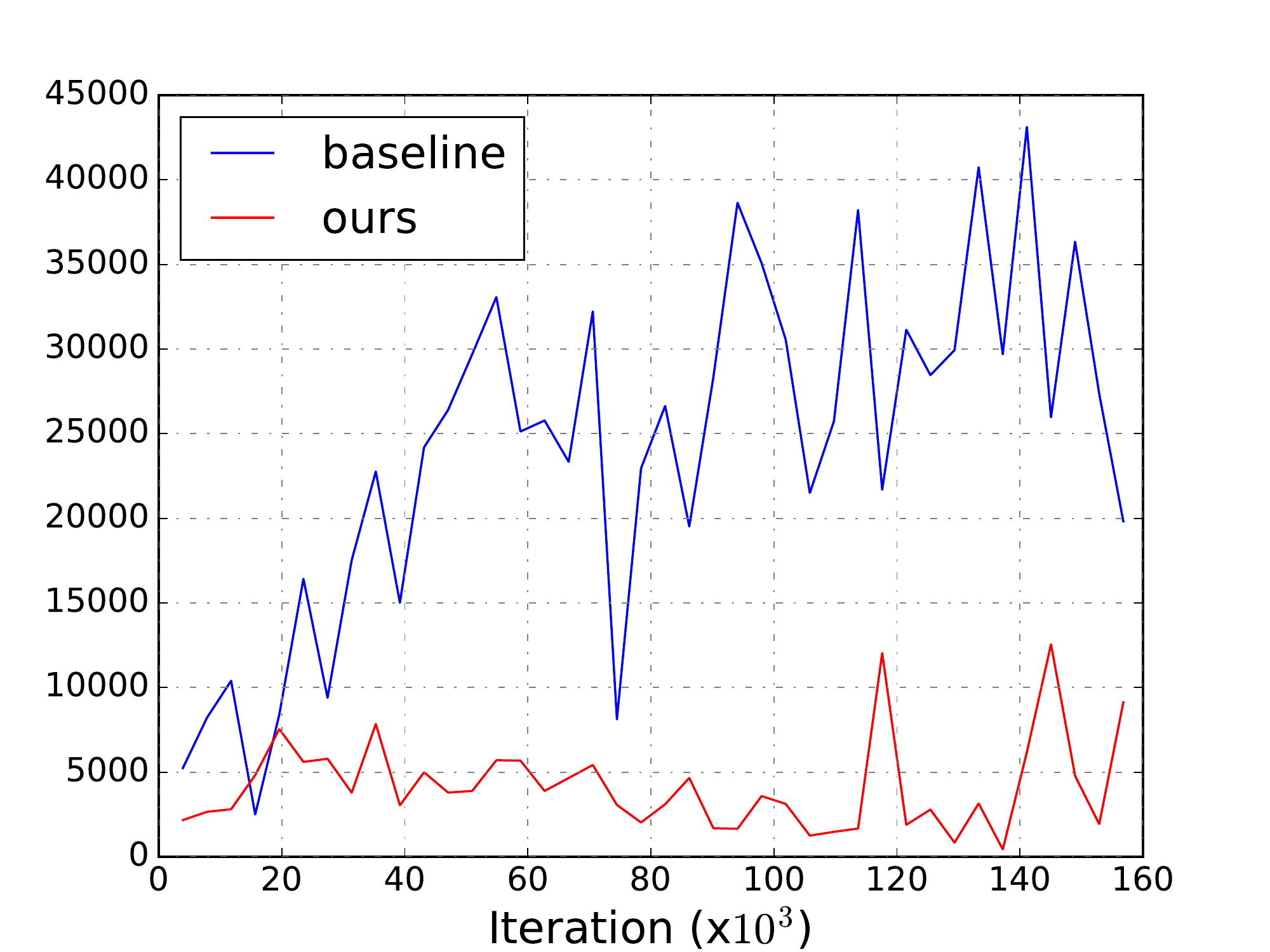}
    }
    \caption{(a) Histogram of the gradient of the discriminator with respect to input images. The left is the baseline, and the right is the adversarially trained discriminator. (b) L1-norm evolution of the discriminator gradient during training. The gradient of the adversarially trained discriminator is sparser and contains less non-informative noise. }
    \label{Grad-ana}
\end{figure}

The gradient given by the discriminator is the key to update the generator. As discussed in Figure \ref{gradient_visual}, the gradient of the adversarially trained discriminator contains some perceptually relevant features, but the gradient of the standard discriminator has no salient pattern that aligns well with human vision. We further give the histogram of the gradient of the discriminator with respect to the real samples as shown in Figure \ref{Grad-ana}a. As the number of training iterations increases, our method can obtain sparser gradients and lower L1 norm (Figure \ref{Grad-ana}b) , which means the adversarial noise in the gradient is partly eliminated. Through a balanced adversarial training on both real and fake samples, the training scheme becomes symmetrical and stable.

\subsection{Training loss}
\label{strategy}
\begin{figure}[H]
    \centering
    \subfigure[]{
        \includegraphics[width=0.45\textwidth]{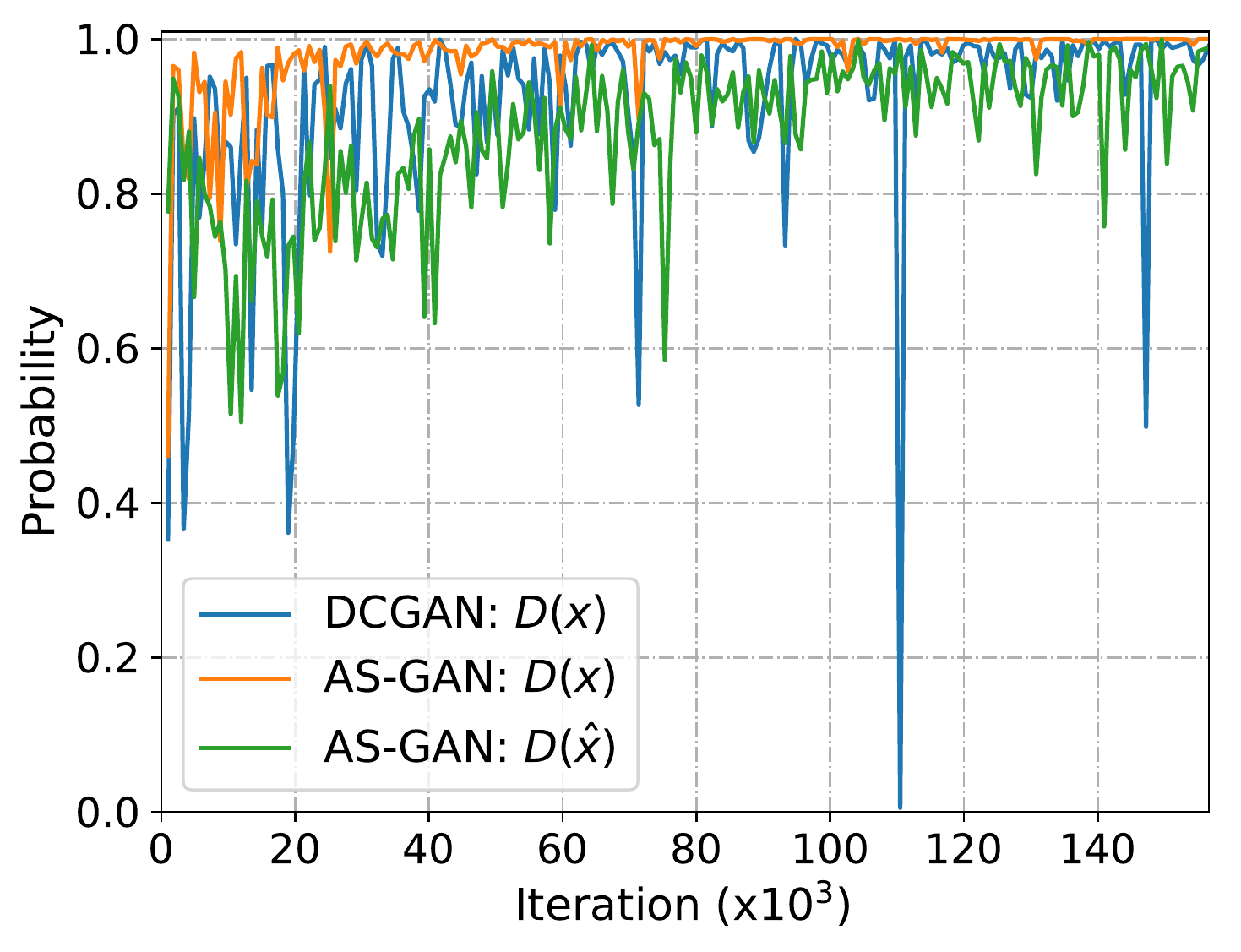}
    }
    \subfigure[]{
        \includegraphics[width=0.45\textwidth]{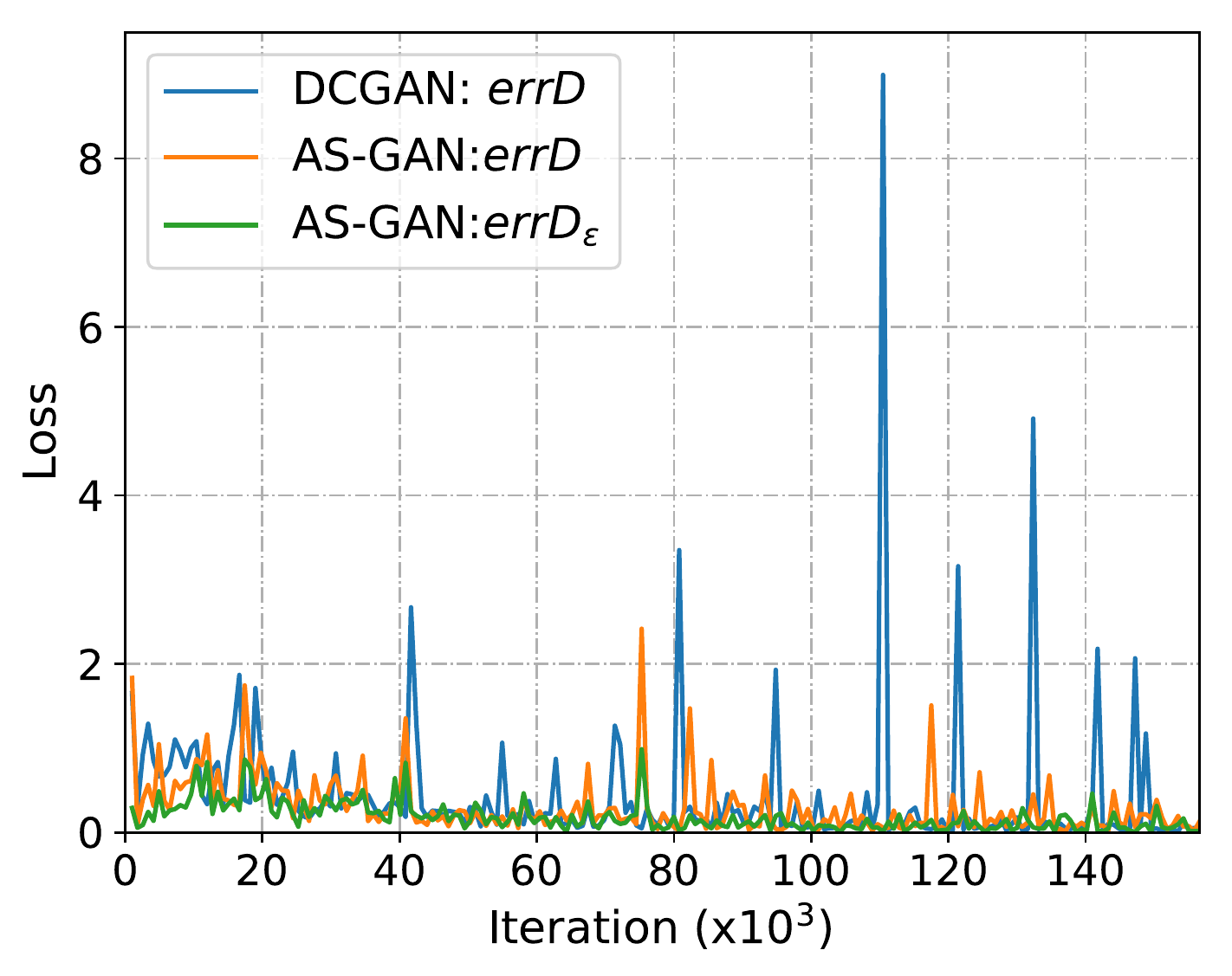}
    }
    \caption{(a) Confidence of the discriminator on real data and adversarial samples of real data during training. (b) Evolution of loss of the discriminator in different settings.}
    \label{proba-loss}
\end{figure}

Using AS-GANs, the confidence and the loss of discriminator become smoother and more stable during training as depicted in Figure \ref{proba-loss}. Moreover, when the adversarial perturbation $\varepsilon$ is appropriate for training, the confidence of adversarial samples $D_{\theta}(\hat{x})$ is much lower than $D_{\theta}(x)$ due to the sensitivity of the discriminator to adversarial samples in the beginning. With more iterations, the discriminator becomes more robust to adversarial samples. As shown in Table \ref{robustness_comparison}, the accuracy on real samples under FGSM attack with perturbation of 1/255 is improved significantly compared with the baseline. Similarly, the loss is also stabilized with our algorithm (Figure \ref{proba-loss}b).

\begin{table}[h]
    \centering
    \begin{threeparttable}
        \caption{Accuracy on real samples under FGSM attack with perturbation of 1/255}
        \label{robustness_comparison}
        \begin{tabular}{ccc}
            \toprule
            Model          & standard accuracy & adversarial accuracy \\
            \midrule
            GAN(ResNet)    & 0.98              & 0.50                 \\
            AS-GAN(ResNet) & 0.99              & 0.93                 \\
            \bottomrule
        \end{tabular}
    \end{threeparttable}
\end{table}

\subsection{Computation overhead}
We have also evaluated the training computation overhead of AS-GANs. An about 25\% overhead relative to the baseline is required, which is comparable to that of spectral normalization but much less than that of gradient penalty. Comparison of average training time of one epoch of different methods is shown in Table \ref{time}.
\renewcommand{\arraystretch}{0.8} 
\begin{table}[h]

    \centering
    \begin{threeparttable}
        \caption{Average training time of different methods}
        \label{time}
        \begin{tabular}{ccccc}
            \toprule
            Setting       & DCGAN  & (ours)AS-DCGAN & SN-DCGAN & DCGAN-GP \\
            \midrule
            Training time & 19.83s & 26.40s         & 24.50s   & 31.57s   \\
            \bottomrule
        \end{tabular}
    \end{threeparttable}
\end{table}

\section{Conclusion}
The relationship between GANs and adversarial samples has remained as an open question since the emergence of these two models. We reveal that adversarial training on fake samples is implemented in standard GANs, but adversarial training on real samples has long been overlooked, making the gradients given by the discriminator contain considerable adversarial noise, which misleads the update of the generator and leads to unstable training. In this work, we incorporate adversarial training on real samples together with fake samples to make training scheme symmetrical and the discriminator more robust. The validation on image generation on CIFAR-10, CIFAR-100, CelebA, and LSUN with varied network architectures demonstrates that adversarial noises can be largely eliminated, thereby significantly improving the convergence speed, performance, and alleviating mode collapse. Besides, AS-GAN with more complex adversarial manipulation methods such as PGD can obtain further improvement. Both intuitive justifications and theoretical analysis are provided to explain why AS-GAN can improve GAN training. Moreover, combining with AS-GANs and spectral normalization, a simple DCGAN can achieve FID comparable to the state-of-the-art on these datasets.
\section{Acknowledgements}
This work is partially supported by the Project of NSFC No. 61836004 and the Brain-Science Special Program of Beijing under Grant Z181100001518006.

\section*{References}

\newpage
\begin{center}
    \textbf{Conflict of Interest}
\end{center}
The authors declared that they have no conflicts of interest to this work.\\

\end{document}